\documentclass[conference]{IEEEtran}
\IEEEoverridecommandlockouts
\usepackage{cite}
\usepackage{amsmath,amssymb,amsfonts}
\usepackage{algorithmic}
\usepackage{graphicx}
\usepackage{textcomp}
\usepackage{xcolor}
\usepackage{tabularx}
\usepackage{booktabs}
\usepackage[export]{adjustbox}
\usepackage{tcolorbox}
\usepackage{hyperref}
\def\BibTeX{{\rm B\kern-.05em{\sc i\kern-.025em b}\kern-.08em
    T\kern-.1667em\lower.7ex\hbox{E}\kern-.125emX}}
\begin{document}

\title{Generating Prototypes for Contradiction Detection Using Large Language Models and Linguistic Rules [Vision Paper]\\

%\author{\IEEEauthorblockN{1\textsuperscript{st} Maren Pielka}
%\IEEEauthorblockA{\textit{Fraunhofer IAIS \\
%Schloss Birlinghoven \\
%53757 Sankt Augustin Germany \\
%Maren.Pielka@iais.fraunhofer.de}}
%\and
%\IEEEauthorblockN{2\textsuperscript{nd} Svetlana Schmidt}
%\IEEEauthorblockA{\textit{Ruhr-Universität Bochum \\
%Universitätsstraße 150 \\ 
%44801 Bochum, Germany \\
%Svetlana.Schmidt@ruhr-uni-bochum.de}}
%\and
%\IEEEauthorblockN{3\textsuperscript{rd} Rafet Sifa}
%\IEEEauthorblockA{\textit{Rheinische Friedrich-Wilhelms Universität Bonn \\
%Regina-Pacis-Weg 3 \\
%53113 Bonn, Germany }}
\author{
    \IEEEauthorblockN{ Maren Pielka\IEEEauthorrefmark{1}\IEEEauthorrefmark{4}, Svetlana Schmidt\IEEEauthorrefmark{1}\IEEEauthorrefmark{2}\IEEEauthorrefmark{4} Rafet Sifa\IEEEauthorrefmark{1}\IEEEauthorrefmark{3}
    }
    \IEEEauthorblockA{\IEEEauthorrefmark{1}Fraunhofer IAIS, Sankt Augustin, Germany}
    \IEEEauthorblockA{\IEEEauthorrefmark{2}Ruhr-Universität Bochum, Bochum, Germany}\IEEEauthorblockA{\IEEEauthorrefmark{3}University of Bonn, Bonn, Germany}
    \IEEEauthorblockA{\IEEEauthorrefmark{4}Equal contribution}
    \texttt{Maren.Pielka@iais.fraunhofer.de}
    \\ 
}

%\and
%\and
%\and
}

\maketitle

\begin{abstract}
We introduce a novel data generation method for contradiction detection, which leverages the generative power of large language models as well as linguistic rules. Our vision is to provide a condensed corpus of prototypical contradictions, allowing for in-depth linguistic analysis as well as efficient language model fine-tuning. To this end, we instruct the generative models to create contradicting statements with respect to descriptions of specific contradiction types. In addition, the model is also instructed to come up with completely new contradiction typologies. As an auxiliary approach, we use linguistic rules to construct simple contradictions such as those arising from negation, antonymy and numeric mismatch. We find that our methods yield promising results in terms of coherence and variety of the data. Further studies, as well as manual refinement are necessary to make use of this data in a machine learning setup.

\end{abstract}


\section{Introduction}
Detecting contradictions in text is one of the hardest tasks for a language model to comprehend. This is due to the complex semantic nature of contradictions, and the variety of contexts in which they can occur. For this reason, a multitude of data sets and models have been developed to solve this task. Meanwhile, the recent onset of large generative language models has given rise to new possibilities for problem solving as well as data augmentation, which we aim to explore in this work.

Contradiction Detection (CD) is a subtask of Natural Language Inference (NLI), but has also been investigated independently in recent years. The goal is, given two pieces of text (premise and hypothesis) that are assumed to refer to the same fact or event, to predict whether there is a contradiction between those. To this end, a contradiction is defined as a mismatch between two statements, such that they cannot possibly be true at the same time. There is of course some subjectivity involved in judging whether two statements are strictly contradicting, or just slightly deviating. Also, the context in which the statements occur can play a crucial role. %For the sake of simplicity, we disregard both of those facts for now, considering only very simple sentences without any context.

Our goal is to build a data set for Contradiction Detection (CD) that conveys different prototypes of contradictions. The idea is to "condense" the essential linguistic properties of contradictions into a relatively small data set, thereby reducing the computational resources needed to train models for solving this task. A side-goal of our work is to extend the typology by \cite{Marneffe2008-ACL} to include more fine-grained contradiction types. We employ an automated method for generating contradictions using rules and large language models (LLMs), which is easily scalable. Our approach for generating the data set is three-fold:

\begin{enumerate}
    \item We generate samples in a rule-based manner by exploiting semantic knowledge graphs and syntactic parsing.
    \item We instruct a large generative language model to produce contradicting hypotheses, given premises from the standard NLI corpus, SNLI \cite{Bowman2015-EMNLP}.
    \item We instruct a large generative language model to produce completely new contradicting statements (both premise and hypothesis), as well as new types of contradictions.
\end{enumerate}

The intuition is to use linguistic and factual rules where this is applicable, i.e. for contradictions based on antonymy, negations and numeric mismatches. For more complex relations such as factive or structural contradictions, we instruct a generative model to produce new samples, either based on given premises or on the type description alone. To our knowledge, this is the first work implementing such a hybrid data generation method with respect to NLI. Our vision is to provide a comprehensive data set, as well as a method to generate more data without much effort, and to broaden the understanding of the complex linguistic nature of contradictions. The code for this paper has been published on Github\footnote{\url{https://github.com/fraunhofer-iais/informed\_nlu/}}.

%\newpage

\section{Related Work}
%todo: add general info about NLI? 1-2 sentences
%todo: describe SNLI dataset and paper
The Stanford Natural Language Inference (SNLI) corpus is a freely available data set which contains 570K sentence pairs \cite{Bowman2015-EMNLP}. The labeled sentence pairs were written by humans based on image capturing. The data were collected with help of the Amazon Mechanical Turk\footnote{\url{https://www.mturk.com/}}. The human workers were asked to provide a hypothesis for a premise scene description. The hypotheses should entail, contradict, or be neutral toward the pre-existing premise \cite{Bowman2015-EMNLP}. 

The approach for creating more realistic data for NLI was presented by \cite{haim2006second}. The texts for premises were collected from news articles. The hypotheses were generated in several ways, using information and relation extraction, question answering, and summarization systems.
%todo:describe de Marneffe Paper

The more fine-grained system for detection of different types of textual contradictions was proposed by \cite{Marneffe2008-ACL}. Following the reasoning in \cite{Marneffe2008-ACL}, there are two main categories of contradictions: 1) the contradictions which arise from antonymy, negation, and numerical mismatch, and 2) the contradictions occurring from subtle lexical differences, contrasting structure of the sentences, factive mismatch, and contrast in the world knowledge (WK). It is quite difficult to detect the contradictions arising from the second category automatically. The understanding of such types is based on the understanding of the sentence meaning.

% idea to generate prototypical examples: Laura's paper

%LLM papers: GPT-3/4, Chain-of-Thought, Self-Instruct
Large (generative) language models (LLMs) have sparked great interest in recent years. Especially the Generative Pretrained Transformer (GPT) framework \cite{radford2018language,radford2019language,NEURIPS2020_1457c0d6} from OpenAI has gained significant popularity - even outside the AI community - since the release of the ChatGPT conversational interface in late 2022. Their latest model GPT-4 \cite{OpenAI23} has set a new state of the art for many language understanding tasks, showing the capability to solve a broad range of real-world tasks such as academic exams on par with human performance. Nevertheless, there are still some shortcomings with respect to those models, for example the fact that they tend to produce incorrect output when asked about complicated or unknown events. Also, they require an extensive amount of data for pre-training, as well as powerful computing resources both for training and inference. There has been some work on utilizing LLMs to create new training data and problem descriptions. Wang et al. \cite{selfinstruct} introduce "Self-Instruct", a framework which can be used to extend the language understanding capabilities of LLMs by having it produce instructions and training instances for language understanding tasks. Those generated instances can then be used to train the LLM further. Our approach is mainly based on that idea, but we focus on the more specific task of detecting contradictions using a linguistic typology.

%pre-training informed with prior knowledge 
The idea of training language models with prototypical knowledge is inspired by \cite{von2022informed}, who suggested to use a similar approach in image classification. They argue that the condensed knowledge of a target domain can be represented by a relatively small data set, which contains training samples that are typical manifestations of the task at hand.
%The novel method of informed pre-training of the neural networks was introduced by \cite{von2022informed}. They proposed to pre-train the machine learning models on prior knowledge which is represented by knowledge prototypes. Those prototypes contain the condensed knowledge of a target domain. For example, the task of recognizing handwritten digits could be represented by a few examples per digit in typical writing styles. The method proposed in \cite{von2022informed} allows to improve the generalization process and to accelerate the pre-training. Our attempt of creating the prototypical contradictions is based on the idea from \cite{von2022informed} that deriving the knowledge prototypes can be applied in different domains.  
%\newpage

There has been some previous work regarding the analysis of the linguistic nature of contradictions, as well as methods to make use of those features in a language modeling setup. \cite{pielka_icmla2022} examine some semantic aspects that are hard to comprehend for machine learning models, such as difference in local prepositions or metaphors. \cite{pielka_ecir2023} build upon those findings by introducing a linguistically informed pre-training regime for encoder-based transformer models, utilizing information about part of speech (POS) tags, synsets and syntactic dependencies. We aim to extend this work by presenting an informed data generation approach which can be used to efficiently fine-tune language models for Contradiction Detection.

\section{Methodology}\label{method}
We argue that the data set for the CD task should include contradictions created in different ways. The generation of contradictions is based on the idea that they arise from different lexical features \cite{Marneffe2008-ACL}. We generate the contradictions based on antonymy, negation, and numerical mismatch in the rule-based approach. The more complex types of contradictions, as described by \cite{Marneffe2008-ACL}, are generated  via the GPT frameworks. 
\subsection{Generating contradictions based on SNLI, using linguistic rules}

For the generation of the hypotheses via rules, we extract syntactic and semantic features from the premises of the SNLI data set. 
The stanza\footnote{\url{https://stanfordnlp.github.io/stanza/depparse.html}} Dependency Parser is utilized for extraction of dependencies, POS tags and morphological features of each sentence.  

For the generation of the antonymy based contradictions we take advantage of the WordNet \cite{miller-1992-wordnet} framework. The content words in WordNet are grouped into synsets (synonym sets) based on their semantic similarity. One meaning of a word is represented by a specific synset.

Contradictions based on antonymy arise when the hypotheses contain antonyms of the aligned words in the premise \cite{Marneffe2008-ACL}. We extract the antonyms of the adjectives and nouns in the premise from WordNet. The hypothesis is then created by replacing the objects and the adjectives of the premise with their antonyms. The meaning of each premise first has to be disambiguated. The disambiguation includes defining one distinct synset for each word in a sentence if available. The SyntagNet API\footnote{\url{http://syntagnet.org/api-documentation}} is utilized for extracting one meaning of each word in a sentence. This step is necessary since there could be several possible synsets for one word. 

We utilize the POS tags, dependencies and morphological features for creating the contradictions based on numerical and polarity mismatch. The contradictory hypotheses are created by negating the verbal phrases of the premises for the contradiction type \textit{negation}. The morphological features are utilized for exerting the right type of the negation, for example the verb phrase in singular number and present tense yields the negative particle \textit{not} together with the modal verb \textit{do}. 

In order to create the numerical mismatch we make use of the dependency \textit{nummod}\footnote{\url{https://universaldependencies.org/u/dep/nummod.html}} (numerical modifier). 
The idea is to create the hypothesis containing numbers bigger or smaller than the ones in the premise.

\subsection{Generating contradictions based on SNLI, using LLMs}
The nature of other contradiction types is more complicated. Generating lexical, structural and factive contradictions, as well as those arising from world knowledge (WK) contrasts, requires more than just changing a pair of words in the hypothesis. The data set collected by \cite{Marneffe2008-ACL} RTE3\footnote{\url{https://nlp.stanford.edu/projects/contradiction/real\_contradiction.xml}} consists of "real$-$life" contradictions which were additionally annotated for their type, and the typology described in their paper is the base for defining the important features of the contradiction types in this work. 
The semantics of the embedding verbs influence the meaning of the whole sentence, and can serve as a basis for entailment or contradiction \cite{nairn2006computing}. According to \cite{Marneffe2008-ACL}, factive contradictions arise from the context in which the verb phrase is embedded, for example \textit{Sudan was ready to accept U.N. troops} contradicts \textit{Sudan refused to accept U.N. troops in Darfur}\footnote{\url{https://nlp.stanford.edu/projects/contradiction/real\_contradiction.xml}}. The opposite meaning of the verb phrase in the hypothesis also creates base for the factive contradiction: 

\begin{center}
    P: Sudan refused to allow U.N. troops in Darfur.\\
    H: Sudan will grant permission for United Nations peacekeeping forces to take up station in Darfur.\footnote{\url{https://nlp.stanford.edu/projects/contradiction/real\_contradiction.xml}}
\end{center}
Structural contradictions arise from the mismatch between the syntactical structures of the premise and hypothesis. This contrast occurs from replacing the object of the verb with the subject from the premise or with a new entity. For example, replacing \textit{parents} in \textit{The children are smiling and waving to their parents} with \textit{friends}: \textit{The children are smiling and waving to their friends} creates contradiction.

As specified by \cite{Marneffe2008-ACL}, lexical contradictions are the type of contradictions which can arise from the distinctive views on an identical event as it is shown in the following example.

\begin{center}
    P: Two women who just had lunch hugging and saying goodbye.
    
    H: The two women who just ate lunch ignored each other and left without saying a word.
\end{center}

The WK contradictions arise from a mismatch in the information regarding one unique event or well-known person \cite{Marneffe2008-ACL}. The example from \cite{Marneffe2008-ACL} illustrates this kind of contradiction:

\begin{center}
    P: President Kennedy was assassinated in Texas.

    H: Kennedy's murder occurred in Washington.
\end{center}

Utilizing LLMs such as GPT allows us to produce a large amount of samples. We use the GPT-4 model with the maximum number of generated tokens set to 512, and the temperature parameter set to 1 for obtaining diverse output. We instruct the GPT-4 model to generate contradictions of each type for every premise and additionally provide it with the descriptions of the different contradiction types and some examples for those contradiction types. The prompt and complete list of the contradiction types' descriptions can be found in the Appendix. 

%An example of the contradiction type description is given below. 

%\begin{tcolorbox}
%    [notitle,boxrule=0pt,boxsep=0pt,left=1em,right=1em,
%top=0.5em,bottom=0.5em,colback=gray!10,colframe=gray!10, fontupper=\color{darkgray}]
%\textbf{Description}:\\Factive contradiction based on the embedding context means that the contradiction arises: 
%\begin{itemize}
%    \item from the mismatch in the embedding context of the verb phrase in the Premise and the Hypothesis
%    \item contains similar or identical entities in the Premise and Hypothesis
%    \item the verb phrase in the Hypothesis have a contradictory meaning
%    \item the Hypothesis does not contain any negations and any antonyms of the verb phrase of the Premise
%\end{itemize}
%\end{tcolorbox}
%TODO: SNLI premises
%Svetlana

\subsection{Generating contradiction types and instances based only on instructions}
%Maren 
As a third approach, we instruct an LLM to generate new instances of specific contradiction types, as well as new typologies. We do not provide the model with any external data, except the descriptions for pre-defined contradiction types, following the typology by \cite{Marneffe2008-ACL}. This approach is inspired by the idea of \cite{selfinstruct}, who proposed to utilize LLMs to jointly generate instructions and instances for language understanding tasks.

\begin{figure}
    \centering
    \includegraphics[width=0.95\linewidth]{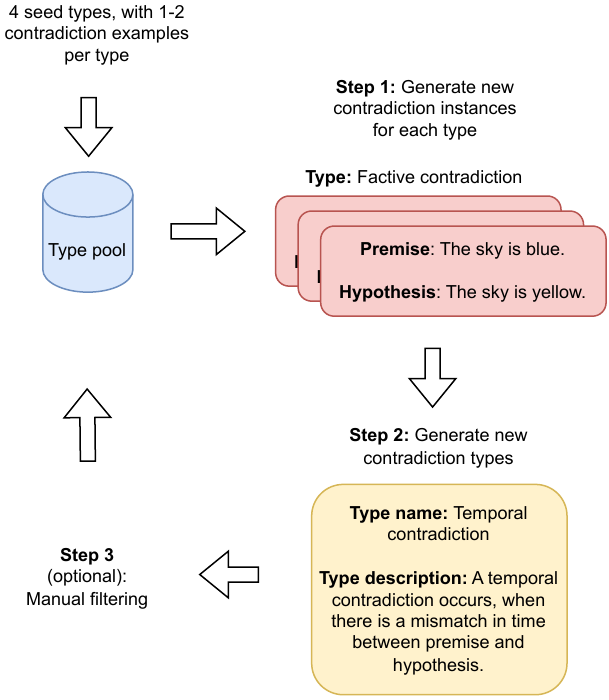}
    \caption{Illustration of the multi-step generation approach.}
    \label{fig:generation}
\end{figure}

The generation process is structured as follows: In every iteration, a fixed number of contradiction samples is being generated for each contradiction type. Additionally, a new type description is being generated, given three randomly sampled descriptions of existing types as examples. The new description is then added to the pool and being used in later iterations for generating both new instances and new types. This policy is depicted in figure \ref{fig:generation}.

We initially start with a pool of contradiction types that contains the classes of contradictions based on structure, lexicality, facts and embedding context as well as verbal antonymy. The prompt templates for generating those contradictions can be found in the Appendix.
For the generation of new contradiction types, GPT-4 is being used. For generating new instances, we experiment with both GPT-3.5 and GPT-4. The parameters of the API call are the same as specified above for the second method.

\section{Theoretical assumptions and preliminary results}
Our main assumption is that a data collection which includes most features of contradictions can help improve the performance of smaller transformer models, like BERT \cite{BERT}, in detecting contradictions. We collected around 3000 samples in total. Among those, 1000 sentence pairs are generated with rules and GPT using SNLI as shown in \ref{tab:data_statistics} (methods 1 and 2). The number of samples generated using Method 3 with GPT only is about 500 sentence pairs (see \ref{tab:data_statistics}). In order to create a balanced data set we add 1500 non-contradiction samples from SNLI. %The relatively small size of the data set can contribute to training the models faster with more computational efficiency. 

\begin{table*}[]
    \centering
    \adjustbox{max width=\textwidth}{\begin{tabular}{lllllllllllll}
        \toprule
            & \multicolumn{3}{c}{Method 1 (rule-based)}  & \multicolumn{4}{c}{Method 2 (GPT with SNLI)}  &\multicolumn{5}{c}{Method 3 (GPT only)}       \\ \cmidrule(lr){2-4} \cmidrule(l){5-8}  \cmidrule(l){9-13} %\midrule
             & Antonymy & Numerical & Negation & Factive & Structure & Lexical & WK &  Factive & Structure & Lexical & WK & Other \\ \midrule
            Number examples & 170 & 165 & 165 & 125 & 125 & 125 & 125 & 50 & 50 & 50 & 50 & 225 \\                         
        \bottomrule 
\end{tabular} } \vspace{1mm}
    \caption{Number of generated examples for the three generation methods, per contradiction type. "Other" stands for all new contradiction types generated by GPT (see Appendix).}
    \label{tab:data_statistics}
\end{table*}

% examples for generated contradictions (good and bad)
The following examples illustrate the contradictions generated with the rule-based approach: 

\begin{tcolorbox}[notitle,boxrule=0pt,boxsep=0pt,left=1em,right=1em,
top=0.5em,bottom=0.5em,colback=gray!10,colframe=gray!10, fontupper=\color{darkgray}]
\textbf{Antonymy}:
\begin{center}
    P: Women exercising one \textit{woman} has a green mat and black outfit on.\\
    H: Women exercising one \textit{man} has a green mat and black outfit on.
\end{center}
\begin{center}
    P: Two \textit{blond} women are hugging one another.\\
    H: Two \textit{brunet} women are hugging one another.
\end{center}
\vspace{2mm}

\textbf{Numerical}:

\begin{center}
    P: \textit{Two} blond women are hugging one another.\\
    H: \textit{Three} blond women are hugging one another.
\end{center}
\vspace{2mm}

\textbf{Negation}:

\begin{center}
    P: Two blond women \textit{are hugging} one another.\\
    H: Two blond women \textit{are not hugging} one another.
\end{center}
\end{tcolorbox}

As it can be seen from the next pair of sentences, the hypothesis created with the rule-based approach  can contain semantic and grammatical errors:

\begin{center}
    P: A \textbf{young} girl sitting at a table with a bowl on her head.\\
    H: A \textbf{old} girl sitting at a table with a bowl on her head.
\end{center}

Thus far, the rules which were used generating contradictory hypotheses are simple. They do not include the semantics of the sentence, and adjustment of the grammatical forms. Nevertheless, the data could still be useful for training a language model.

Here are some examples for hypotheses which have been generated using GPT-4, given the respective premise:

\begin{tcolorbox}[notitle,boxrule=0pt,boxsep=0pt,left=1em,right=1em,
top=0.5em,bottom=0.5em,colback=gray!10,colframe=gray!10, fontupper=\color{darkgray}]
\textbf{Factive}:

\begin{center}
    P: Children are smiling and waving at the camera.\\
    H: Children are crying and ignoring the camera.
\end{center}

\begin{center}
    P: Children are smiling and waving at the camera.\\
    H: Children forgot to smile and wave at the camera.
\end{center}
%\vspace{2mm}
\end{tcolorbox}

\begin{tcolorbox}[notitle,boxrule=0pt,boxsep=0pt,left=1em,right=1em,
top=0.5em,bottom=0.5em,colback=gray!10,colframe=gray!10, fontupper=\color{darkgray}]

\textbf{Structure}:

\begin{center}
    P: A couple is playing with a little boy on the beach.\\
    H: A couple is playing with a dog on the beach.
\end{center}
\vspace{2mm}

\textbf{Lexical}:

\begin{center}
    P: A boy is jumping on skateboard in the middle of a red bridge.\end{center}
\begin{center}
    H: The kid is sitting while riding his bike at the end of a green passage.
\end{center}
\vspace{2mm}

\textbf{World Knowledge}:

\begin{center}
    P: A person on a horse jumps over a broken down airplane.\\
    H: Airplanes are too large and tall for a horse to jump over.
\end{center}

\end{tcolorbox}

One of the difficulties of the contradiction generation with the GPT model is the limited number of sentences it can generate at one time. Another complexity in generating the contradictions is that GPT models can produce sentences which contain semantic errors or do not correspond to the world knowledge, as it is illustrated in the following example:

\begin{center}
    P: A person on a horse jumps over a broken down airplane.\\
    H: The broken down airplane overleaps the person on a horse.
\end{center}

With respect to the third method, where we instruct the LLM to produce both premise and hypothesis given a contradiction type description, there are significant differences in quality between the contradiction types. For some cases the generation works reasonably well, as can be seen in this example of a lexical contradiction:

\begin{center}
    P: The cat is sleeping peacefully on the couch.\\
    H: The cat is wide awake and running around the room.
\end{center}

In other cases - specifically for structural contradictions - the approach does not work well at all, as the language model produces grammatically correct, but semantically meaningless hypotheses:

\begin{center}
    P: He cooked delicious pasta for dinner.\\
    H: The pasta cooked him deliciously for dinner.
\end{center}

Surprisingly, switching from GPT-3.5 to GPT-4 for the instance generation does not change the quality of the results much.

As described in section \ref{method}, we also instruct GPT-4 to generate completely new types of contradictions. This works surprisingly well, as can be seen in this example (both the type description, as well as the instance have been generated by GPT-4):

\begin{tcolorbox}[notitle,boxrule=0pt,boxsep=0pt,left=1em,right=1em,
top=0.5em,bottom=0.5em,colback=gray!10,colframe=gray!10, fontupper=\color{darkgray}]
    \textbf{Temporal mismatch} \vspace{1mm} \\ 
    \textbf{Description:} This contradiction arises when there's inconsistency between the time frames or chronological\end{tcolorbox}

\begin{tcolorbox}[notitle,boxrule=0pt,boxsep=0pt,left=1em,right=1em,
top=0.5em,bottom=0.5em,colback=gray!10,colframe=gray!10, fontupper=\color{darkgray}]events presented in two statements. Hypothetically, if one statement indicates an event happening before another, and the contradictory statement implies the opposite sequence or suggests the events are simultaneous, a temporal mismatch is present. \vspace{1mm} \\ 
    \textbf{Example:} \\
    P: The movie was released two months ago. \\
    H: The actors are currently filming the sequel.

\end{tcolorbox}

Most of the types produced by GPT are logically coherent and semantically meaningful. The LLM also manages to generate reasonable instances for each of those types. We observe that in some cases the same type of contradiction is effectively generated multiple times, even though the description varies slightly (e.g. for "Temporal mismatch").
Nevertheless, those new types of contradictions could possibly contribute to better understanding the underlying semantics, and offer a more fine-grained typology. A complete list of all newly generated types (after duplication removal) is provided in the Appendix.
% Todo Maren: results from method 3
% examples for new contradiction types
% challenges (output parsing)
% (maybe) first quantitative results
%\newpage

\section{Conclusion and Outlook}
In this work we combined two approaches for data generation: a rule-based and an LLM based approach. Preliminary results have shown that simple contradiction types, such as contradictions based on antonymy, negation and numerical mismatch can be generated using simple constraints. The advantage of this approach is that the contradictions created in this way contain the major features of the contradiction types, e.g. the aligned pair of words being antonyms. %Thus, they represent the contradiction prototypes. 
The generation of contradictions with the GPT models allows to create more complex contradiction types, such as factive, structural, lexical, and those based on world knowledge. The most important part while creating those contradiction types is describing the subtle characteristics of each type. Our first attempts in instructing the GPT model have shown that LLMs can comprehend the instructions and create meaningful contradictions according to specific contradiction types descriptions. 

%Using those two methods we were able to produce the required number of contradictions.
One advantage of our method is the purposeful generation of specific types of contradictions. Assuming that particular types of contradictions prevail in specific domains, only the relevant contradiction types could be generated. For example, it is \cite{deusser2023contradiction} emphasizes that contradictions occurring in the financial reports are more often of numeric and lexical nature.

As it was stated in section \ref{method}, some samples display semantic and grammatical errors. One possible solution could be manual filtering, meaning the generated data could be additionally validated and possibly refined by human annotators. %After this optimization step the prototypes of the contradictions will be utilized for pre-training the language models on the CD task.
We plan to address this in our future work.

There is a variety of possible use cases for the proposed method. It could e.g. be used to build a data set for fake news detection, or identifying contradictory statements in financial reports. To this end, it would be meaningful to have the data set contain longer paragraphs to represent the real-world scenario better. We also plan to provide more tailored instructions and data for specific applications.

\section*{Acknowledgment}
This research has been partially funded by the Federal Ministry of Education and Research of Germany and the state of North-Rhine Westphalia as part of the Lamarr-Institute for Machine Learning and Artificial Intelligence.

\bibliography{bibliography}
\bibliographystyle{plain}
\newpage

\section{Appendix}
\label{appendix}
\subsection{Prompts for generating contradiction instances and types}
The following prompt is utilized for generating the hypotheses for the list of premises:

\begin{tcolorbox}[notitle,boxrule=0pt,boxsep=0pt,left=1em,right=1em,
top=0.5em,bottom=0.5em,colback=gray!10,colframe=gray!10, fontupper=\color{darkgray}]
\textbf{System}: You are an expert on semantics and linguistics, with a profound knowledge in Natural Language Processing. You are especially aware of the work by Marneffe et al., classifying different types of contradictions, such as factive, structural, lexical and world knowledge contradictions. The Premise is provided, you have to create a Hypothesis of one of the contradiction types for this premise.
\textbf{User:} Please generate one contradictory Hypothesis for a PREMISE, based on CONTRADICTION\_TYPE\_DESCRIPTION. Format your response in the following way: CONTRADICTION\_TYPE\_NAME 'P: [PREMISE], H: [HYPOTHESIS]'.
\textbf{Assistant:} CONTRADICTION\_TYPE\_DESCRIPTION
\end{tcolorbox}

The placeholders stand for the following entities:
\begin{itemize}
    \item PREMISE: the premise from SNLI data set which is used by the model as base for hypothesis generation
    \item CONTRADICTION\_TYPE\_NAME: Name of the type of contradiction that should be generated
    \item CONTRADICTION\_TYPE\_DESCRIPTION: Short description of the contradiction type to generate.
\end{itemize}

We use the following prompt for generating new contradiction instances, given a specific contradiction type: 

\begin{tcolorbox}[notitle,boxrule=0pt,boxsep=0pt,left=1em,right=1em,
top=0.5em,bottom=0.5em,colback=gray!10,colframe=gray!10, fontupper=\color{darkgray}]
\textbf{System}: You are an expert on semantics and linguistics, with a profound knowledge in Natural Language Processing. You are especially aware of the work by Marneffe et al., classifying different types of contradictions, such as contradictions arising from antonymy, negation, or numeric mismatch. To this end, a contradiction is defined as a mismatch between two statements, such that they cannot possibly both be true. It is assumed, that both statements refer to the same fact or event, even if this is not explicitly stated. \vspace{2mm} \\ 
\textbf{User:} Please generate NUM\_CONTRADICTIONS different contradictions based on CONTRADICTION\_TYPE\_NAME. The contradictions should be original and reasonably different from each other. Both premise and hypothesis should contain at least 10 words each, and should not be too similar. Please take care that they are actually contradicting and semantically meaningful. Be creative! Format your response in the following way: 'Premise: [PREMISE], Hypothesis: [HYPOTHESIS]'. Keep to this format strictly and do not add extra text or numbers.\vspace{2mm} \\ 
\textbf{Assistant}: CONTRADICTION\_TYPE\_DESCRIPTION
\end{tcolorbox}

The placeholders stand for the following entities:
\begin{itemize}
    \item NUM\_CONTRADICTIONS: Pre-defined number of contradictions to generate per type and iteration (set to 5)
    \item CONTRADICTION\_TYPE\_NAME: Name of the type of contradiction that should be generated
    \item CONTRADICTION\_TYPE\_DESCRIPTION: Short description of the contradiction type to generate.
\end{itemize}

The following prompt is being used for generating new contradiction types:

\begin{tcolorbox}[notitle,boxrule=0pt,boxsep=0pt,left=1em,right=1em,
top=0.5em,bottom=0.5em,colback=gray!10,colframe=gray!10, fontupper=\color{darkgray}]
\textbf{System}: You are an expert on semantics and linguistics, with a profound knowledge in Natural Language Processing. You are especially aware of the work by Marneffe et al., classifying different types of contradictions, such as contradictions arising from antonymy, negation, or numeric mismatch.\vspace{2mm} \\ 
\textbf{User:} Please come up with a new category of contradiction (other than KNOWN\_TYPES). Format your output in the following way: Contradiction type name: [TYPE\_NAME], Contradiction type description: [TYPE\_DESCRIPTION].\vspace{2mm} \\ 
\textbf{Assistant}: CONTRADICTION\_TYPE\_DESCRIPTIONS
\end{tcolorbox}

Here the placeholders stand for:

\begin{itemize}
    \item KNOWN\_TYPES: List of all contradiction types already known at that point (both initial and self-generated).
    \item CONTRADICTION\_TYPE\_DESCRIPTIONS: List of descriptions for three randomly selected contradiction types from the pool of existing types.
\end{itemize}

\subsection{Descriptions of the contradiction types which were used in prompts for contradiction generation}

\begin{tcolorbox}[notitle,boxrule=0pt,boxsep=0pt,left=1em,right=1em,
top=0.5em,bottom=0.5em,colback=gray!10,colframe=gray!10, fontupper=\color{darkgray}]
\textbf{Factive (embedding context)}:
Factive contradiction based on the embedding context means that a contradiction:
\begin{itemize}
    \item arises from the mismatch in the embedding context of the verb phrase in the Premise and Hypothesis;
    creates the contradictory meaning;
    \item contains similar or identical entities in the Premise and Hypothesis;
    \item Hypothesis does not contain any negations and any antonyms of the verb phrase of the Premise.
\end{itemize}
\textbf{Example}:\\
P: Sudan accepted U.N. troops in Darfur.\\
H: Sudan refused to accept U.N. troops.\footnote{\url{https://nlp.stanford.edu/projects/contradiction/real\_contradiction.xml}}
\end{tcolorbox}

\begin{tcolorbox}[notitle,boxrule=0pt,boxsep=0pt,left=1em,right=1em,
top=0.5em,bottom=0.5em,colback=gray!10,colframe=gray!10, fontupper=\color{darkgray}]
\textbf{Factive (antonymy based)}:
Factive contradiction based on the antonymy of a verb means that a contradiction arises between two statements (Premise and Hypothesis), because the verb phrase in Hypothesis has an opposite or contradictory meaning      compared to the verb phrase in the Premise.
 \end{tcolorbox}
    \begin{tcolorbox}[notitle,boxrule=0pt,boxsep=0pt,left=1em,right=1em,
top=0.5em,bottom=0.5em,colback=gray!10,colframe=gray!10, fontupper=\color{darkgray}]
\textbf{Example}:\\
P: Sudan refused to allow U.N. troops in Darfur.\\
H: Sudan will grant permission for United Nations peacekeeping forces to take up station in Darfur.\footnote{\url{https://nlp.stanford.edu/projects/contradiction/real\_contradiction.xml}}
\end{tcolorbox}

\begin{tcolorbox}[notitle,boxrule=0pt,boxsep=0pt,left=1em,right=1em,
top=0.5em,bottom=0.5em,colback=gray!10,colframe=gray!10, fontupper=\color{darkgray}]
\textbf{Structure}: 
Structure contradiction arises
from the mismatch in the sentence structure of the premise and hypothesis. The mismatch in the sentence structure has following features:
\begin{itemize}
    \item the created Hypothesis has the same verb phrase as the Premise;
    \item there are new entities which function as new objects of the same verb in the hypothesis, which creates the contradictory meaning toward the meaning of the premise 
\end{itemize}
\textbf{Example}:\\
P: The children are smiling and waving at the camera.
H: The children are smiling and waving to each other.
\end{tcolorbox}

\begin{tcolorbox}[notitle,boxrule=0pt,boxsep=0pt,left=1em,right=1em,
top=0.5em,bottom=0.5em,colback=gray!10,colframe=gray!10, fontupper=\color{darkgray}]
\textbf{Lexical}:
Lexical contradiction based on the mismatch in the lexical context has following features:
\begin{itemize}
    \item the Premise and Hypothesis has both the same topic or verb subject
    \item the created Hypothesis has subtly different lexical meaning
    \item the Hypothesis has a contradictory meaning due to the created opposite context of the topic in the premise
\end{itemize}
\textbf{Example}:\\
P: Tariq Aziz kept outside the closed circle of Saddam s Sunni Moslem cronies.\\
H: Tariq Aziz was in Saddam s inner circle.\footnote{\url{https://nlp.stanford.edu/projects/contradiction/real\_contradiction.xml}}
\end{tcolorbox}

\begin{tcolorbox}[notitle,boxrule=0pt,boxsep=0pt,left=1em,right=1em,
top=0.5em,bottom=0.5em,colback=gray!10,colframe=gray!10, fontupper=\color{darkgray}]
\textbf{World Knowledge}
Lexical contradiction based on the mismatch in world knowledge has following features:
\begin{itemize}
    \item the Premise contains the well known knowledge about the world
    \item the facts and knowledge from the Hypothesis contradict to the world knowledge in the Premise
\end{itemize}
Examples: Premise='Al-zarqawi was Palestinian.'
Hypothesis='Al-zarqawi was Jordanian.'\footnote{\url{https://nlp.stanford.edu/projects/contradiction/real\_contradiction.xml}}

\end{tcolorbox}
\newpage

\subsection{Descriptions of the contradiction types generated by GPT-4}
% To Do Maren
\begin{tcolorbox}[notitle,boxrule=0pt,boxsep=0pt,left=1em,right=1em,
top=0.5em,bottom=0.5em,colback=gray!10,colframe=gray!10, fontupper=\color{darkgray}]
    \textbf{Temporal mismatch} \vspace{1mm} \\ 
    \textbf{Description:} This contradiction arises when there's inconsistency between the time frames or chronological events presented in two statements. Hypothetically, if one statement indicates an event happening before another, and the contradictory statement implies the opposite sequence or suggests the events are simultaneous, a temporal mismatch is present. \vspace{1mm} \\ 
    \textbf{Example:} \\
    P: The movie was released two months ago. \\
    H: The actors are currently filming the sequel.

\end{tcolorbox}

\begin{tcolorbox}[notitle,boxrule=0pt,boxsep=0pt,left=1em,right=1em,
top=0.5em,bottom=0.5em,colback=gray!10,colframe=gray!10, fontupper=\color{darkgray}]
    \textbf{Aspectual contradictions} \vspace{1mm} \\ 
    \textbf{Description:} These contradictions arise from mismatches in the aspectual properties of verbs or verb phrases between the premise and the claim. Aspect refers to the temporal structure of events or states as they are viewed from a specific standpoint in time. Aspectual contradictions can occur when the same event-state is characterized in conflicting ways; for example, in terms of its completion, frequency, duration or temporality. For instance, a premise stating 'John has been running for an hour' and a claim asserting 'John just started running' would form an aspectual contradiction, as they present the same action but with incompatible aspectual properties; specifically, contradictory assertions about the action's duration or initiation. \vspace{1mm} \\ 
    \textbf{Example:} \\
    P: Mary has been studying French for years. \\
    H: Mary has never studied French before.

\end{tcolorbox}

\begin{tcolorbox}[notitle,boxrule=0pt,boxsep=0pt,left=1em,right=1em,
top=0.5em,bottom=0.5em,colback=gray!10,colframe=gray!10, fontupper=\color{darkgray}]
    \textbf{Causal mismatch} \vspace{1mm} \\ 
    \textbf{Description:} This contradiction arises when the cause and effect relationship implied in one statement is fundamentally at odds with or invalidated by another statement. For example, if one statement posits that a certain result is due to a specific cause, and a contradictory statement suggests that the same result is due to a completely different cause, or that the first cause doesn't lead to the mentioned effect, a causal mismatch is present. The contradiction is formed because the cause-and-effect relationships in the statements are incompatible. For example, a premise saying 'Rain makes the road slippery' and a claim stating 'Rain makes the road dry' would constitute a causal mismatch. \vspace{1mm} \\ 
    \textbf{Example:} \\
    P: Eating a healthy diet leads to weight loss. \\
    H: Eating junk food leads to weight loss.

\end{tcolorbox}

\begin{tcolorbox}[notitle,boxrule=0pt,boxsep=0pt,left=1em,right=1em,
top=0.5em,bottom=0.5em,colback=gray!10,colframe=gray!10, fontupper=\color{darkgray}]
    \textbf{Spatial mismatch} \vspace{1mm} \\ 
    \textbf{Description:} This type of contradiction occurs when two statements or pieces of information present conflicting descriptions of physical or spatial arrangements.
    For example, one might state that a certain object or person is in a specific location, while the other places it somewhere else. This includes contradictions related to proximity, relative position, direction and geographical location.
\vspace{1mm}  \\
    \textbf{Example:} \\
    P: The house is located on the top of the hill. \\
    H: The house is situated in a valley deep underground.

\end{tcolorbox}

\begin{tcolorbox}[notitle,boxrule=0pt,boxsep=0pt,left=1em,right=1em,
top=0.5em,bottom=0.5em,colback=gray!10,colframe=gray!10, fontupper=\color{darkgray}]
    \textbf{Ideological mismatch} \vspace{1mm} \\ 
    \textbf{Description:} This type of contradiction arises when two statements, while not necessarily directly opposing, conflict based on underlying ideological, philosophical, or theoretical frameworks. This could involve contradictions originating from differences in belief systems, moral values, or personal convictions. These contradictions may not result from antonymy, negation, or numeric mismatch.
    Rather, they emanate from deeper cognitive dissonance or juxtaposition of incongruent worldviews. For instance, two statements like 'Justice is swift punishment' and 'Justice is rehabilitation not punishment' may constitute an ideological mismatch, as they are based on fundamentally different beliefs about what 'justice' entails. \vspace{1mm} \\ 
    \textbf{Example:} \\
    P: Capitalism is the only economic model that promotes and preserves individual liberty. \\
    H: Socialism is a beneficial economic model that supports collective welfare and liberty.
\end{tcolorbox}

\begin{tcolorbox}[notitle,boxrule=0pt,boxsep=0pt,left=1em,right=1em,
top=0.5em,bottom=0.5em,colback=gray!10,colframe=gray!10, fontupper=\color{darkgray}]
    \textbf{Modal mismatch} \vspace{1mm} \\ 
    \textbf{Description:} This category of contradiction arises when two statements are discordant in the modalities they imply or express. Modalities can range from possibility, necessity, obligation, permission, and ability. For example, the premise may assert that a course of action is necessary, while the contradicting statement may imply that the same action is merely possible or even unnecessary. This mismatch in modal claims leads to contradiction. \vspace{1mm} \\ 
    \textbf{Example:} \\
    P: John is legally obliged to finish the project by next week as stated in their contractual agreement. \\
    H: John has the option to complete the project anytime he wishes without any mandatory deadlines.
\end{tcolorbox}

\begin{tcolorbox}[notitle,boxrule=0pt,boxsep=0pt,left=1em,right=1em,
top=0.5em,bottom=0.5em,colback=gray!10,colframe=gray!10, fontupper=\color{darkgray}]
    \textbf{Quantitative mismatch} \vspace{1mm} \\ 
    \textbf{Description:} This type of contradiction arises when two statements conflict due to inconsistent quantitative information. \end{tcolorbox}
    \begin{tcolorbox}[notitle,boxrule=0pt,boxsep=0pt,left=1em,right=1em,
top=0.5em,bottom=0.5em,colback=gray!10,colframe=gray!10, fontupper=\color{darkgray}]Unlike numeric mismatch where explicit numbers contradict each other, quantitative mismatch happens when imprecise measures, orders of magnitude, or qualitative quantities clash between statements. For example, one statement might refer to an event or entity as being 'rare', while a conflicting statement describes it as 'common'. Similarly, one could say 'He consumed a large amount of water', while another says, 'He had little to drink.' This category is subtle as it requires inferencing and understanding of relative measures and estimates to detect contradictions. \vspace{1mm} \\ 
    \textbf{Example:} \\
    P: The attendance at the local football match was exceptionally high, filling the stadium to the brim. \\
    H: The local football match was not popular, with most of the stadium remaining empty.
\end{tcolorbox}

\begin{tcolorbox}[notitle,boxrule=0pt,boxsep=0pt,left=1em,right=1em,
top=0.5em,bottom=0.5em,colback=gray!10,colframe=gray!10, fontupper=\color{darkgray}]
    \textbf{Probabilistic mismatch} \vspace{1mm} \\ 
    \textbf{Description:} This type of contradiction occurs when two statements provide different estimations of probability or likelihood for the same event or outcome. One statement may suggest that something is very likely to happen, while the other statement asserts that it's very unlikely or even impossible. In a broader sense, the contradiction can also include cases where the level of certainty or definiteness implicated in the statements is at odds. For instance, 'John will certainly attend the party' versus 'It's unlikely that John will attend the party' represents a Probabilistic Mismatch. \vspace{1mm} \\ 
    \textbf{Example:} \\
    P: The meteorologist stated with certainty that the hurricane will strike the coast tomorrow morning. \\
    H: The weather report forecasts a small chance of the hurricane reaching the coast tomorrow morning.
\end{tcolorbox}

\end{document}